\documentclass[11pt,a4paper]{article}
\usepackage[hyperref]{acl2021}
\usepackage{times}
\usepackage{latexsym}

\newcommand{\myfont}{\fontsize{10pt}{\baselineskip}\selectfont}
\newcommand{\myfontnew}{\fontsize{9.5pt}{\baselineskip}\selectfont}
\usepackage{graphicx}
\usepackage{amsmath}
\usepackage{CJKutf8}
\DeclareMathOperator*{\argmax}{arg\,max}

\usepackage{amsthm}
\usepackage{amssymb}
\usepackage{booktabs}
\usepackage{algorithm}
\usepackage{array}
\usepackage{subfigure}
\usepackage{multirow}
\usepackage{threeparttable}
\usepackage{tabularx}
\usepackage{booktabs}
\usepackage{colortbl}
\usepackage{xcolor}
\usepackage{stfloats}
\usepackage[noend]{algpseudocode}
\usepackage{mathtools}
\usepackage{makecell}
\usepackage{setspace}
\usepackage{microtype}
\aclfinalcopy 

\newcolumntype{M}[1]{>{\centering\arraybackslash}m{#1}}
\title{SENT: Sentence-level Distant Relation Extraction via Negative Training}

 \author{
    Ruotian Ma$^{1}$, Tao Gui$^{2}$\thanks{$^*$  Corresponding authors.} , Linyang Li$^{1}$, Qi Zhang$^{1*}$, Yaqian Zhou$^{1}$ and Xuanjing Huang$^{1}$  \\
  $^1$School of Computer Science, Fudan University, Shanghai, China \\
  $^2$Institute of Modern Languages and Linguistics, Fudan University, Shanghai, China \\
  \texttt{\{rtma19,tgui16,linyangli19,qz,yqzhou,xjhuang\}@fudan.edu.cn}
  }
\date{}

\begin{document}
\maketitle
\begin{abstract}
Distant supervision for relation extraction provides uniform bag labels for each sentence inside the bag, while accurate sentence labels are important for downstream applications that need the exact relation type. Directly using bag labels for sentence-level training will introduce much noise, thus severely degrading performance. In this work, we propose the use of negative training (NT), in which a model is trained using complementary labels regarding that ``the instance does not belong to these complementary labels". Since the probability of selecting a true label as a complementary label is low, NT provides less noisy information. Furthermore, the model trained with NT is able to separate the noisy data from the training data. 
Based on NT, we propose a sentence-level framework, SENT, for distant relation extraction. SENT not only filters the noisy data to construct a cleaner dataset, but also performs a re-labeling process to transform the noisy data into useful training data, thus further 
benefiting the model's performance. Experimental results show the significant improvement of the proposed method over previous methods on sentence-level evaluation and de-noise effect.

\end{abstract}

\section{Introduction}\label{intro}

Relation extraction (RE), which aims to extract the relation between entity pairs from unstructured text, is a fundamental task in natural language processing. The extracted relation facts can benefit various downstream applications, e.g., knowledge graph completion  \cite{bordes2013translating,wang2014knowledge}, 
information extraction \cite{wu-weld-2010-open}
and question answering \cite{yao-van-durme-2014-information,fader2014open}. 

A significant challenge for relation extraction is the lack of large-scale labeled data. Thus, distant supervision \cite{mintz2009distant} is proposed to gather training data through automatic alignment between a database and plain text. Such annotation paradigm results in an inevitable noise problem, which is alleviated by previous studies using multi-instance learning (MIL). In MIL, the training and testing processes are performed at the bag level, where a bag contains noisy sentences mentioning the same entity pair but possibly not describing the same relation. Studies using MIL can be broadly classified into two categories: 1) the soft de-noise methods that leverage soft weights to differentiate the influence of each sentence \cite{lin2016neural,han2018hierarchical,Li_Long_Shen_Zhou_Yao_Huo_Jiang_2020,hu-etal-2019-improving,ye2019distant,yuan2019distant,yuan2019cross}; 2) the hard de-noise methods that remove noisy sentences from the bag \cite{zeng2015distant,qin2018robust,han2018denoising,shang2019noisy}.

\begin{figure}
    \centering
    \includegraphics[width=1.0\linewidth]{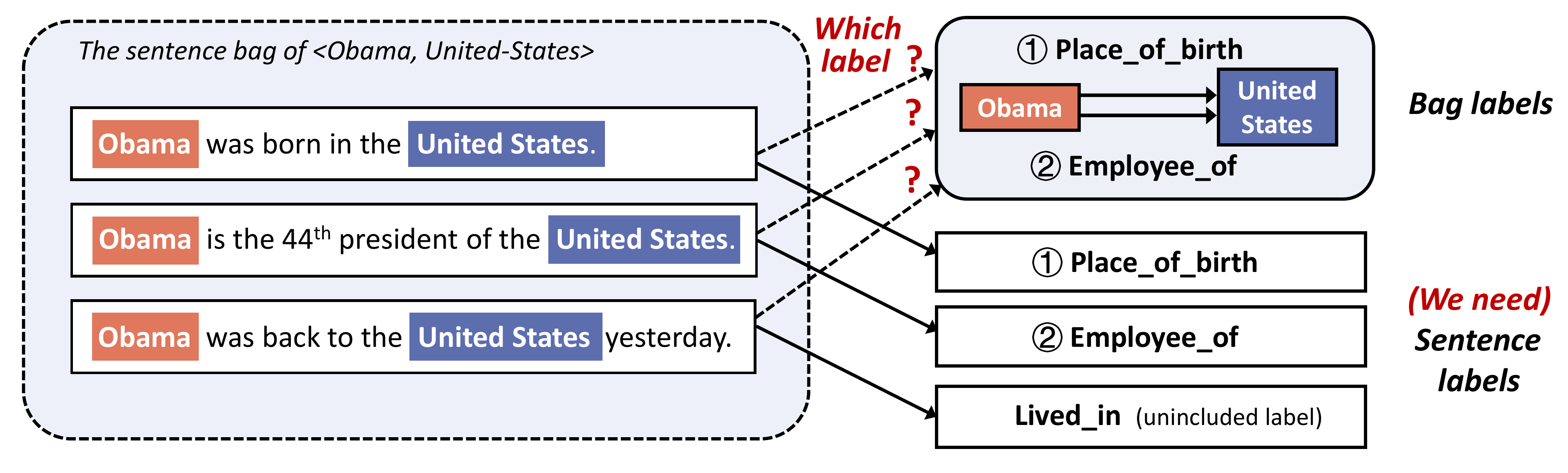}
    \caption{Two types of noise exist in bag-level labels: 1) Multi-label noise: the exact label (``place\_of\_birth" or ``employee\_of") for each sentence is unclear; 2) Wrong-label noise: the third sentence inside the bag actually expresses ``live\_in" which is not included in the bag labels. }
    \label{fig:baglabel}
    \vspace{-0.2cm}
\end{figure}

However, these bag-level approaches fail to map each sentence inside bags with explicit sentence labels. This problem limits the application of RE in some downstream tasks that require sentence-level relation type, e.g., \citet{yao-van-durme-2014-information} and \citet{ xu-etal-2016-question} use sentence-level relation extraction to identify the relation between
the answer and the entity in the question. Therefore, several studies (\citet{jia-etal-2019-arnor,feng2018reinforcement}) have made efforts on sentence-level (or instance-level) distant RE, empirically verifying the deficiency of bag-level methods on sentence-level evaluation. However, the instance selection approaches of these methods depend on rewards\cite{feng2018reinforcement} or frequent patterns\cite{jia-etal-2019-arnor} determined by bag-level labels, which contain much noise. For one thing, one bag might be assigned to multiple bag labels, leading to difficulties in one-to-one mapping between sentences and labels. As shown in Fig.\ref{fig:baglabel}, we have no access to the exact relation between ``place\_of\_birth" and ``employee\_of" for the sentence ``Obama was born in the United States.". For another, the sentences inside a bag might not express the bag relations. In Fig.\ref{fig:baglabel}, the sentence ``Obama was back to the United States yesterday" actually express the relation ``live\_in", which is not included in the bag labels.

In this work, we propose the use of negative training (NT) \cite{kim2019nlnl} for distant RE.
Different from positive training (PT), NT trains a model by selecting the complementary labels of the given label, regarding that ``the input sentence does not belong to this complementary label". Since the probability of selecting a true label as a complementary label is low, NT decreases the risk of providing noisy information and prevents the model from overfitting the noisy data.
Moreover, the model trained with NT is able to separate the noisy data from the training data (a histogram in Fig.\ref{fig:distribution} shows the separated data distribution during NT). Based on NT, we propose SENT, a sentence-level framework for distant RE. During SENT training, the noisy instances are not only filtered with a noise-filtering strategy, but also transformed into useful training data with a re-labeling method. We further design an iterative training algorithm to take full advantage of these data-refining processes, which significantly boost performance. Our codes are publicly available at \textit{Github}\footnote{https://github.com/rtmaww/SENT}.

To summarize the contribution of this work:
\begin{itemize}
\setlength{\itemindent}{0em}
\setlength{\itemsep}{0em}
\setlength{\topsep}{-0.5em}
    \item We propose the use of negative training for sentence-level distant RE, which greatly protects the model from noisy information.
    \item We present a sentence-level framework, SENT, 
    which includes a noise-filtering and a re-labeling strategy for re-fining distant data. 
    \item The proposed method achieves significant improvement over previous methods in terms of both RE performance and de-noise effect.
\end{itemize}

\section{Related Work}
\subsection{Distant Supervision for RE}
Supervised relation extraction (RE) has been constrained by the lack of large-scale labeled data. Therefore, distant supervision (DS) is introduced by \citet{mintz2009distant}, which employs existing knowledge bases (KBs) as source of supervision instead of annotated text. 
\citet{riedel2010modeling} relaxes the DS assumption to the express-at-least-once assumption. As a result, multi-instance learning is introduced (\citet{riedel2010modeling, hoffmann-etal-2011-knowledge, surdeanu-etal-2012-multi}) for this task, where the training and evaluating process are performed in \textbf{bag-level}, with potential noisy sentences existing in each bag. Most following studies in distant RE adopt this paradigm, aiming to decrease the impact of noisy sentences in each bag.  These studies include the attention-based methods to attend to useful information ( \citet{lin2016neural,han2018hierarchical,Li_Long_Shen_Zhou_Yao_Huo_Jiang_2020,hu-etal-2019-improving,ye2019distant,yuan2019distant,zhu2019improving,yuan2019cross,wu-etal-2017-adversarial}), the selection strategies such as RL or adversarial training to remove noisy sentences from the bag (\citet{zeng2015distant,shang2019noisy,qin2018robust,han2018denoising}) and the incorporation with extra information such as KGs, multi-lingual corpora or other information (\citet{ji2017distant,lei-etal-2018-cooperative,vashishth-etal-2018-reside,han2018neural,zhang2019long,qu2019fine,verga-etal-2016-multilingual,lin-etal-2017-neural,wang-etal-2018-adversarial,deng-sun-2019-leveraging,beltagy-etal-2019-combining}). Other approaches include soft-label strategy for denoising (\citet{liu-etal-2017-soft}), leveraging pre-trained LM (\citet{alt-etal-2019-fine}), pattern-based method (\citet{zheng-etal-2019-diag}), structured learning method (\citet{bai-ritter-2019-structured}) and so forth (\citet{luo-etal-2017-learning,chen-etal-2019-uncover}).

In this work, we focus on \textbf{sentence-level} relation extraction.
Several previous studies also perform Distant RE on sentence-level. \citet{feng2018reinforcement} proposes a reinforcement learning framework for sentence selecting, where the reward is given by the classification scores on bag labels. \citet{jia-etal-2019-arnor}  builds an initial training set and further select confident instances based on selected patterns. The difference between the proposed work and previous works is that we do not rely on bag-level labels for sentence selecting. Furthermore, we leverage NT to dynamically separate the noisy data from the training data, thus can make use of diversified clean data.

\subsection{Learning with Noisy Data}
Learning with noisy data is a widely discussed problem in deep learning, especially in the field of computer vision. Existing approaches include robust learning methods such as leveraging a robust loss function or regularization method\cite{Lyu2020Curriculum,NEURIPS2018_f2925f97,hu2019simple,kim2019nlnl}, re-weighting the loss of potential noisy samples \cite{pmlr-v80-ren18a,pmlr-v80-jiang18c}, modeling the corruption probability with a transition matrix \cite{goldberger2016training,xiaanchor} and so on. Another line of research tries to recognize or even correct the noisy instances from the training data\cite{malach2017decoupling,yu2019does,arazo2019unsupervised,li2019dividemix}. 

In this paper, we focus on the noisy label problem in distant RE. We first leverage a robust negative loss \cite{kim2019nlnl} for model training. Then, we develop a new iterative training algorithm for noise selection and correction. 


\begin{figure*}
    \centering
    \includegraphics[width=1.0\linewidth]{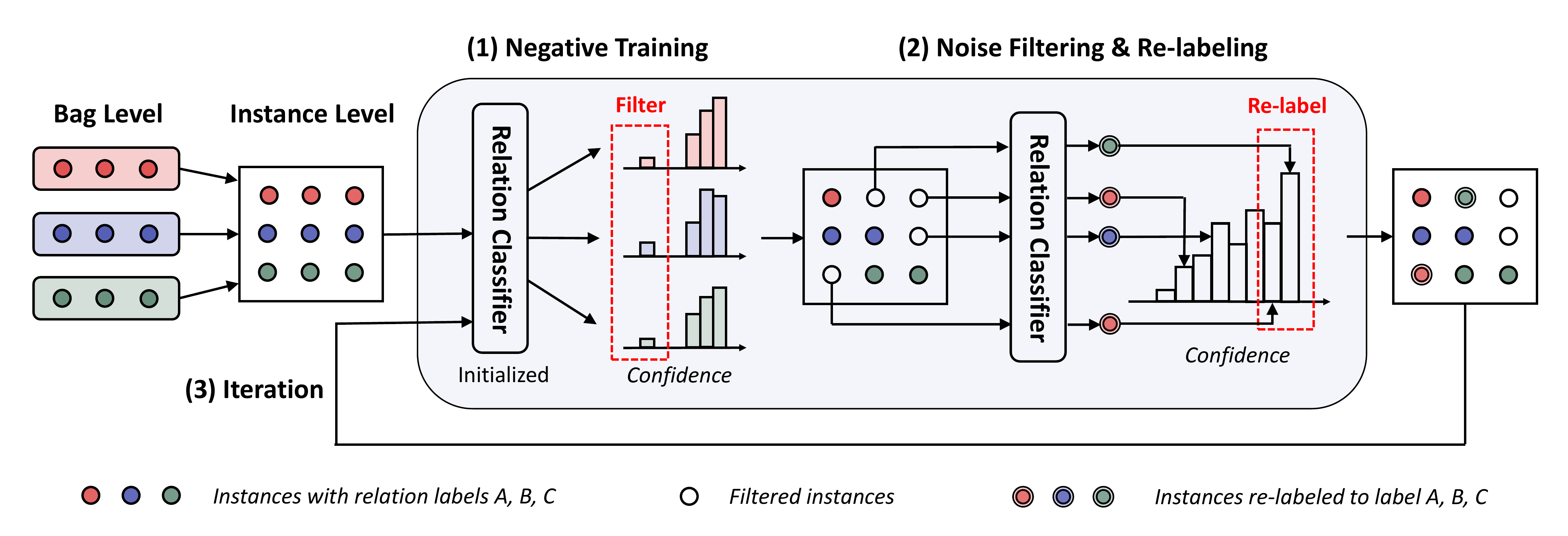}
    \caption{An overview of the proposed framework, SENT, for sentence-level distant RE. Three steps are included: (1) Negative training for separating the noisy data from the training data; (2) Noise-filtering and re-labeling; (3) Iterative training to further boost the performance.}
    \label{fig:algorithm}
\end{figure*}

\section{Methodology}
In order to achieve sentence-level relation classification using bag-level labels in distant RE, we propose a framework, SENT, which contains three main steps (as shown in Fig.\ref{fig:algorithm}): (1) Separating the noisy data from the training data with negative training (Sec.\ref{NT}); (2) Filtering the noisy data as well as re-labeling a part of confident instances (Sec.\ref{filter_relabel}); (3) Leveraging an effective training algorithm based on (1) and (2) to further boost the performance (Sec.\ref{iteration}).

Specifically, we denote the input data in this task as $\mathbf{{S}^*}=\{(s_1, {y}_1^*), \dots, (s_N, {y}_N^*)\}$, where ${y}_i^*\in \mathbb{R}=\{1,\dots, C\}$ is the bag-level label of the $i^{th}$ input sentence $s_i$. Obviously, this is a noisy dataset drawn from a noisy distribution $\mathbf{{D}^*}$ because these bag-level labels ${y}^*$ come from the distant label of each entity bag. For each $s_i$ containing a pair of entities $<e_1, e_2>$, ${y}^*_i$ is one of the relation facts\footnote{Here, we randomly choose one of the multiple bag labels for injective relation classification. See details in Sec.\ref{implementation}.} that $<e_1, e_2>$ participates in in the database. Such annotation method indicates that ${y}^*_i$ is a potential noisy label for $s_i$.
Here, we denote $\mathbf{D}$ as the real data distribution without noise, and the clean dataset drawn from $\mathbf{D}$ as $\mathbf{{S}}=\{(s_1, {y}_1), \dots, (s_N, {y}_N)\}$. The ambition of this work is to find the best estimated parameters $\theta$ of the real mapping $f: x \to y, (x,y)\in \mathbf{D} $ based on the noisy data $\mathbf{{S}^*}$. We design three steps for achieving this goal: (1) Recognizing the set of noisy data $\mathbf{S}^*_n$ from $\mathbf{{S}^*}$ using negative training, where $\mathbf{S}^*_n=\{(s_i,{y}^*_i) \mid {y}^*_i \neq y_i\}$. (2) Refining $\mathbf{{S}^*}$ by noise-filtering and re-labeling, e.g., $\mathbf{S}^*_{refined}=(\mathbf{{S}^*}\setminus \mathbf{S}^*_n) \cup \mathbf{S}^*_{n,relabeled}$, where $\mathbf{S}^*_{n,relabeled}=\{(s_i,{y}_i) \mid (s_i,{y}^*_i) \in \mathbf{S}^*_n \} $. (3) Iteratively perform (1) and (2) so the refined dataset $\mathbf{S}^*_{refined} $ approaches the real dataset $\mathbf{{S}}$.

\begin{figure*}[htbp]
\centering
\subfigure[Positive training]{
\begin{minipage}[t]{0.25\linewidth}
\includegraphics[width=1.0\linewidth]{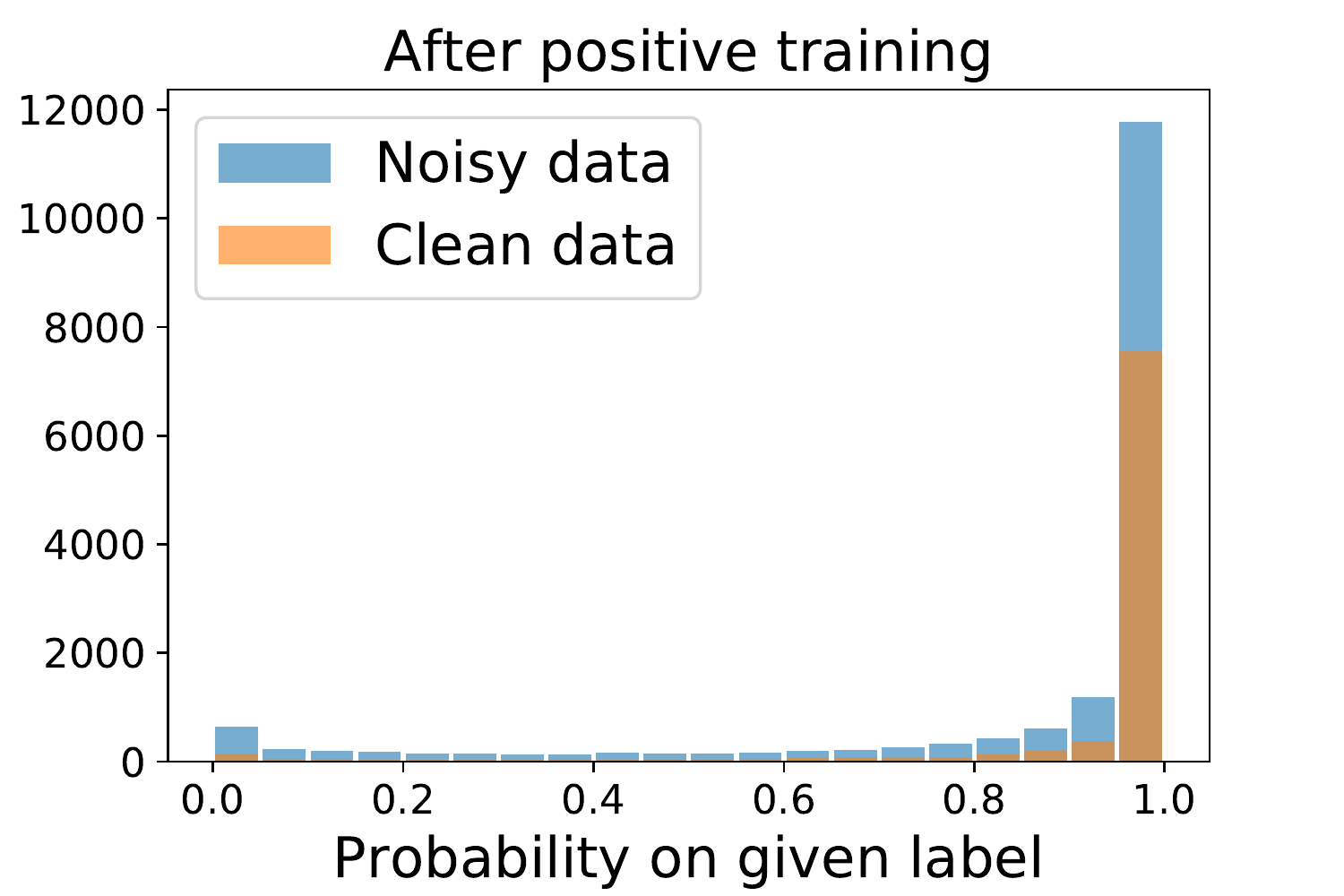}
\end{minipage}%
}%
\subfigure[Negative training]{
\begin{minipage}[t]{0.25\linewidth}
\includegraphics[width=1.0\linewidth]{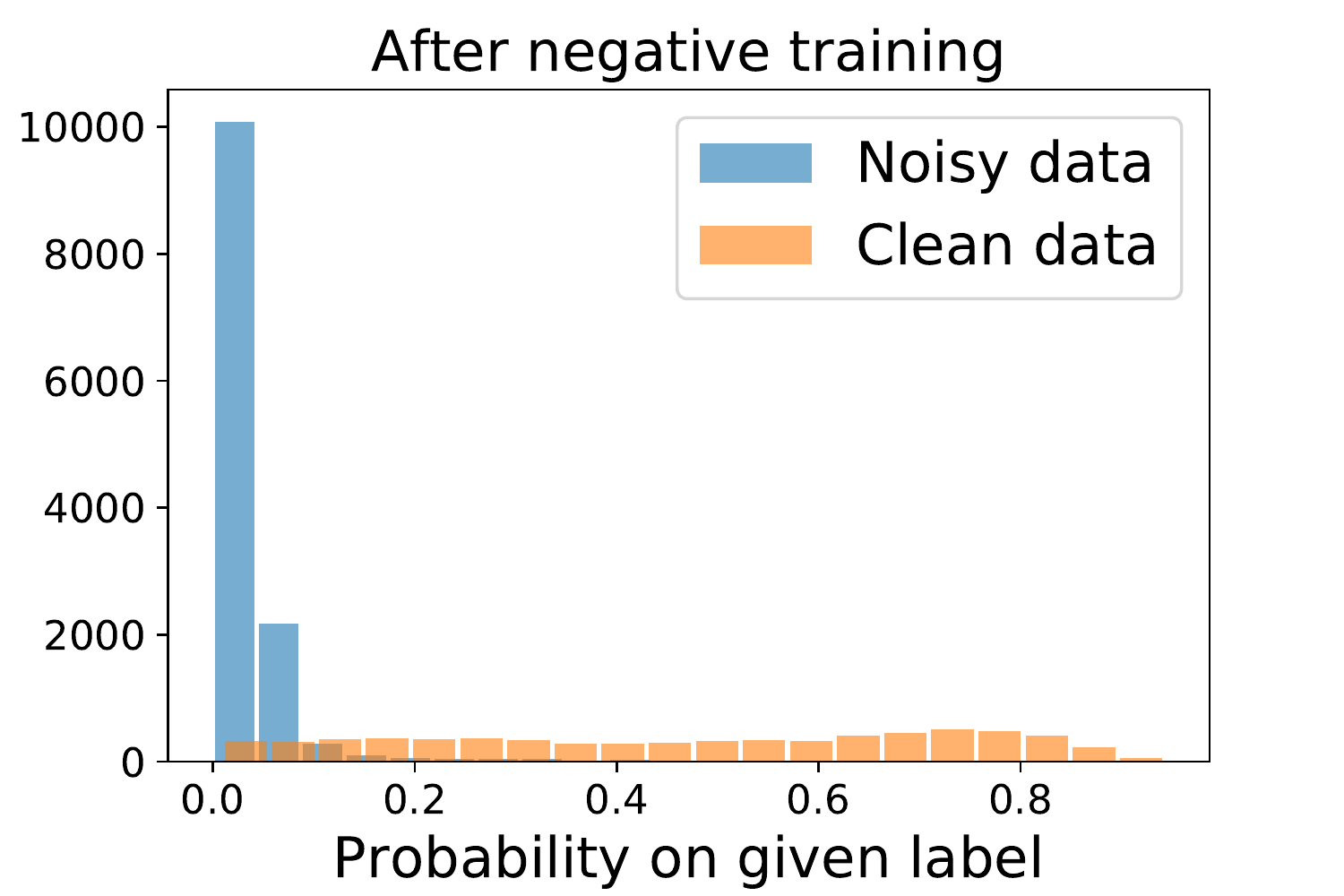}
\end{minipage}%
}%
\subfigure[After SENT]{
\begin{minipage}[t]{0.25\linewidth}
\includegraphics[width=1.0\linewidth]{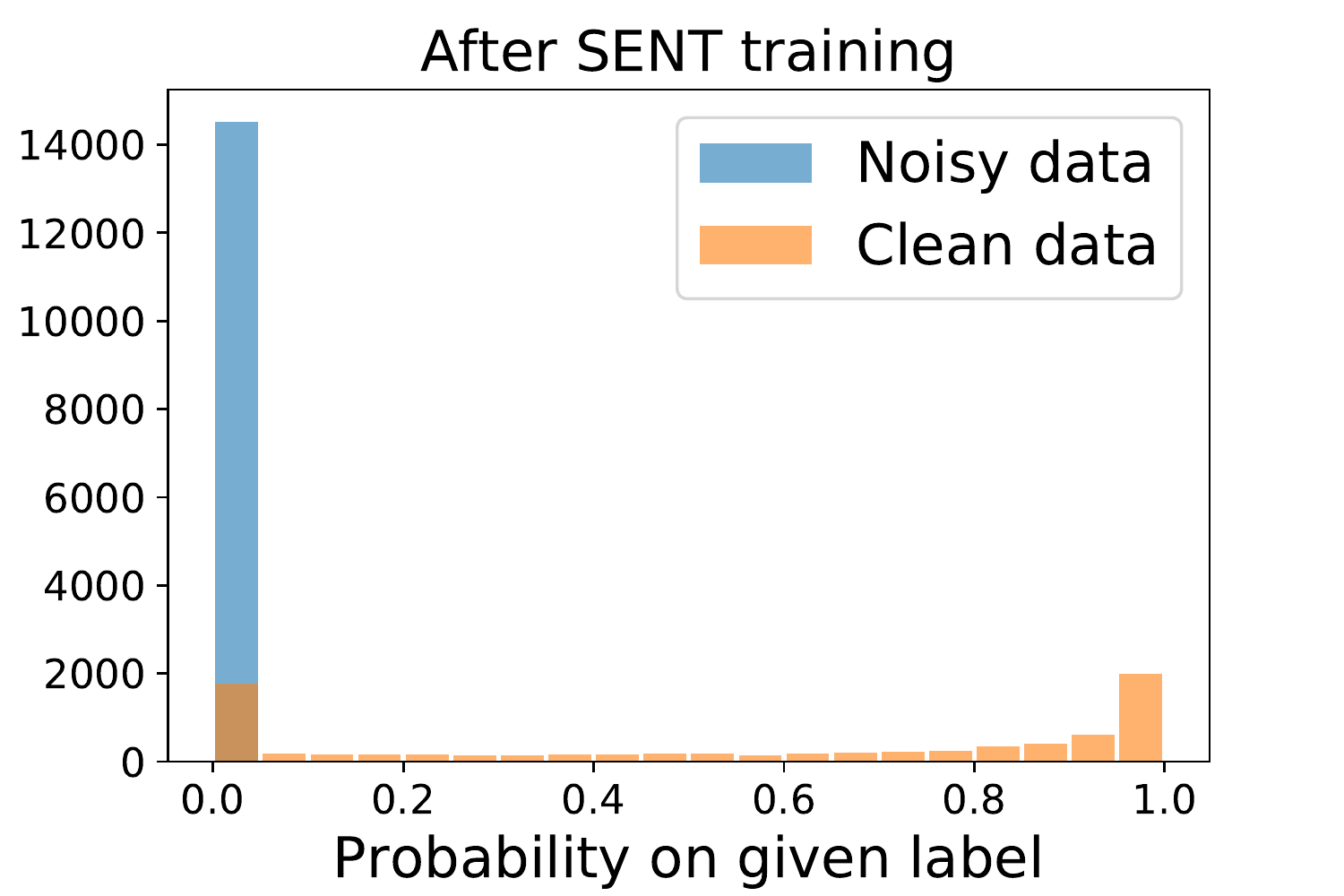}
\end{minipage}%
}%
\subfigure[PT after SENT]{
\begin{minipage}[t]{0.25\linewidth}
\includegraphics[width=1.0\linewidth]{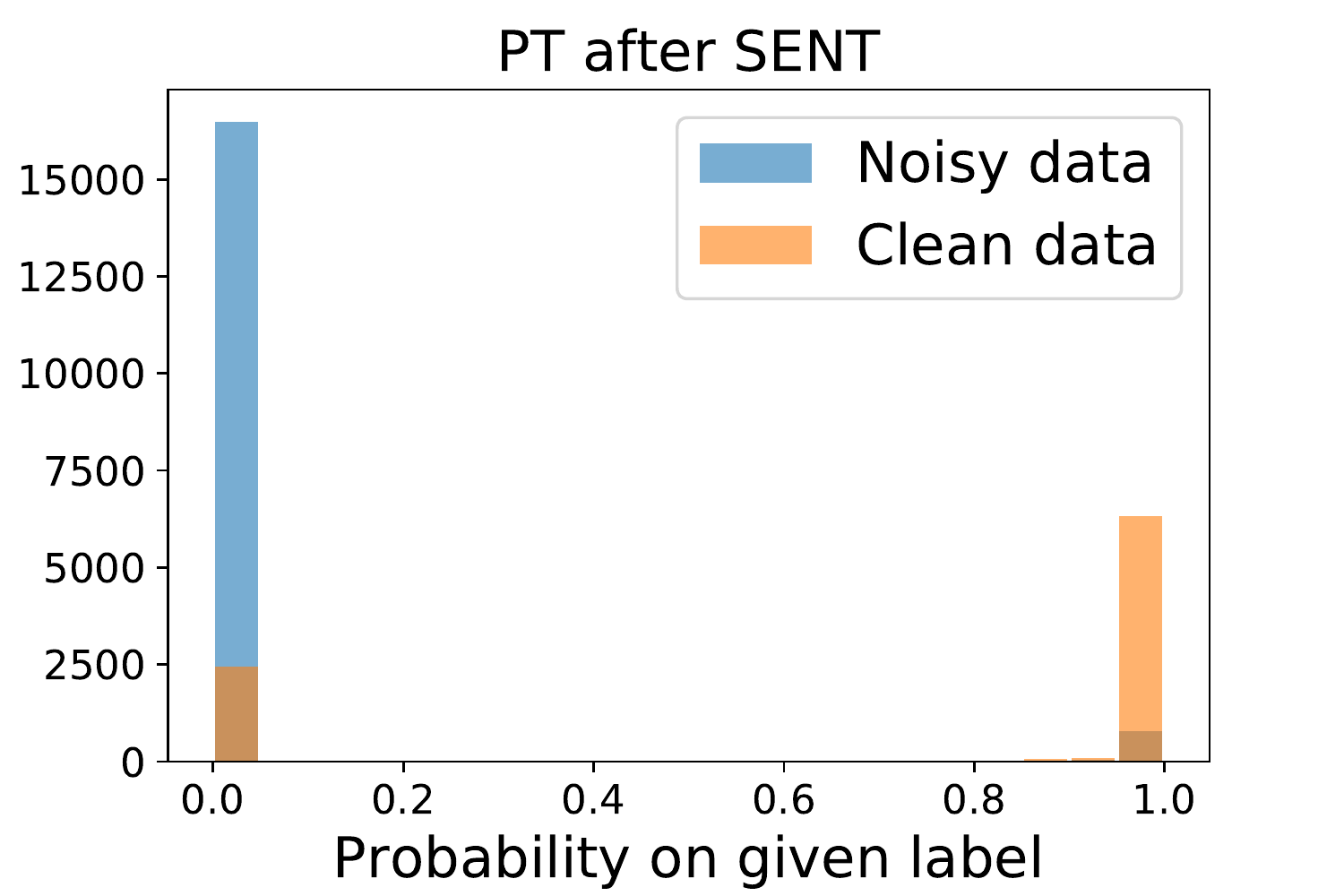}
\end{minipage}%
}%

\centering
\caption{Data distribution when training with PT and SENT. (a) During PT, the confidence of the clean and noisy data increase simultaneously; (b) During NT, the confidence of the noisy data is much lower than that of the clean data; (c) After training with the SENT method, the clean and noisy data are further separated; (d) PT after SENT helps improve the convergence of the clean data.}
\label{fig:distribution}
\end{figure*}

\subsection{Negative Training on Distant Data}\label{NT}
In order to perform robust training on the noisy distant data, we propose the use of negative Training (NT), which trains based on the concept that ``the input sentence does not belong to this complementary label". We find that NT not only provides less noisy information, but also separates the noisy and clean data during training.

\subsubsection{Positive Training}
Positive training (PT) trains the model towards predicting the given label, based on the concept that ``the input sentence belongs to this label". Here, given any input ${s}$ with a label ${y}^* \in \mathbb{R}=\{1, 2, \dots, C\}$, $\textbf{y} \in \{0,1\}^C$ is the C-dimension one-hot vector of ${y}^*$. We denote $\mathbf{p}=f({s})$ as the probability vector of a sentence given by a relation classifier $f(\cdot)$.
With the cross entropy loss function, the loss defined in typical positive training is:
\begin{equation}\label{pl}
    \mathcal{L}_{PT}(f, {y}^*) = -\sum_{k=1}^C {y}_k \log {p}_k
\end{equation}
where ${p}_k$ denotes the probability of the $k^{th}$ label. Optimizing on Eq.\ref{pl} meets the requirement of PL, as the probability of the given label approaches 1 with the loss decreasing. 

\subsubsection{Negative Training}
In negative training (NT), for each input ${s}$ with a label ${y}^* \in \mathbb{R}$, we generate a complementary label $\overline{y^*}$ by randomly sampling from the label space except ${y}^*$, e.g., $\overline{y^*} \in \mathbb{R} \backslash \{{y^*}\}$. With the cross entropy loss function, we define the loss in negative training as:
\begin{equation}\label{nl}
    \mathcal{L}_{NT}(f, {y}^*) = -\sum_{k=1}^C \overline{{y}_k} \log(1- p_k)
\end{equation}
Different from PT, Eq.\ref{nl} aims to reduce the probability value of the complementary label, as $p_k \to 0$ with the loss decreasing. 

To further illustrate the effect of NT, we train the classifier with PT and NT respectively on a constructed TACRED dataset with 30\% noise (details shown in Sec.4.1). A histogram\footnote{When drawing the histogram, we omitted the large amount of ``NA"-class data (80\% of the training data) for a clearer representation of the positive-class data.} of the training data after PT and NT is shown in Figs. \ref{fig:distribution}(a),(b), which reveals that, when training with PT, the confidence of clean data and noisy data increase with no difference, resulting in the model to overfit noisy training data. On the contrary, when training with NT, the confidence of noisy data is much lower than that of clean data. This result confirms that the model trained with NT suffers less from overfitting noisy data with less noisy information provided. Moreover, as the confidence value of clean data and noisy data separate from each other, we are able to filter noisy data with a certain threshold. Fig.\ref{fig:data_refine} shows the details of the data-filtering effect. After the first iteration of NT, a modest threshold contributes to 97\% precision noise-filtering with about 50\% recall, which further verifies the effectiveness of NT on noisy data training.



\subsection{Noise Filtering and Re-labeling}\label{filter_relabel}

In Section \ref{NT}, we have illustrated the effectiveness of NT on training with noisy data, as well as the capability to recognize noisy instances. 
While filtering noisy data is important for training on distant data, these filtered data contain useful information that can boost performance if properly re-labeled.
In this section, we describe the proposed noise-filtering and label-recovering strategy for refining distant data based on NT.


\subsubsection{Filtering Noisy Data}\label{denoising}
As discussed before, it is intuitive to construct a filtering strategy based on a certain threshold after NT. However, in distant RE, the long-tail problem cannot be neglected. During training, the degree of convergence is disparate among different classes. Simply setting a uniform threshold might harm the data distribution with instances of long-tail relations largely filtered out. Therefore, we leverage a dynamic threshold for filtering noisy data. Suppose the probability of class $c$ of the $i^{th}$ instance is $p_c^i \in (0, p_c^h)$, where $p_c^h$ is the maximum probability value in class $c$. Based on empirical experience, we assume the probability values follow a distribution where the noisy data are largely distributed in low-value areas and the clean data are generally distributed in middle- or high-value areas. Therefore, the filtering threshold of class $c$ is set to:
\begin{equation}\label{filter}
    Th_c = Th\cdot p_c^h, p_c^h = \max_{i=1}^N\{p_c^i\}
\end{equation}
where $Th$ is a global threshold. In this way, the noise-filtering threshold not only relies on the degree of convergence in each class, but also dynamically changes during the training phase, thus making it more suitable for noise-filtering on long-tail data. 

\subsubsection{Re-labeling Useful Data}
After noise-filtering, the noisy instances are regarded as unlabeled data, which also contain useful information for training. Here, we design a simple strategy for re-labeling these unlabeled data. Given the set of filtered data $D_u=\{{s}_1, \dots, {s}_m\}$, we use the classifier trained in this iteration to predict the probability vectors $\{ \mathbf{p}^1, \dots, \mathbf{p}^m\}$. Then, we re-label these instances by:
\begin{equation}
    \label{relabel}
    \hat{y}_i = \argmax_k\{p^i_k\}, if \max_k\{p^i_k\} > Th_{relabel}
\end{equation}
where $p^i_k$ is the probability of the $i^{th}$ instance in class k, and $Th_{relabel}$ is the re-label threshold.

\subsection{Iterative Training Algorithm}\label{iteration}

Although effective, simply performing a pipeline of NT, noise-filtering and re-labeling fail to take full advantage of each part, thus the model performance can be further boosted through iterative training.  

As shown in Fig.\ref{fig:algorithm}, for each iteration, we first train the classifier on the noisy data using NT: for each instance, we randomly sample $K$ complementary labels and calculate the loss on these labels with Eq.(\ref{nl}). After $M$-epochs negative training, the noise-filtering and re-labeling processes are carried out for updating the training data. Next, we perform a new iteration of training on the newly-refined data. Here, we re-initialize the classifier in every iteration for two reasons: First, re-initialization ensures that in each iteration, the new classifier is trained on a dataset with higher quality. Second, re-initialization introduces randomness, thus contributing to more robust data-filtering. Finally, we stop the iteration after observing the best result on the dev set. We then perform a round of noise-filtering and re-labeling with the best model in the last iteration to obtain the final refined data.

Fig.\ref{fig:distribution}(c) shows the data distribution after certain iterations of SENT. As seen, the noise and clean data are separated by a large margin. Most noisy data are successfully filtered out, with an acceptable number of clean data mistaken. However, we can see that the model trained with NT still lacks convergence (with low-confidence predictions). Therefore, we train the classifier on the iteratively-refined data with PT for better convergence. As shown in Fig.\ref{fig:distribution}(d), the model predictions on most of the clean data are in high confidence after PT training.


\section{Experiments}
The experiments in this work are divided into two parts, respectively conducted on two datasets: the NYT-10 dataset \cite{riedel2010modeling} and the TACRED dataset \cite{zhang-etal-2017-position}.

The first part is the effectiveness study on sentence-level evaluation for distant RE. Different from bag-level evaluation, a sentence-level evaluation compute Precision (Prec.), Recall (Rec.) and F1 metric directly on all of the individual instances in the dataset.
In this part, we adopt the NYT-10 data set for sentence-level training,  following the setting of \citet{jia-etal-2019-arnor}, who publishes a manually labeled sentence-level test set. \footnote{https://github.com/PaddlePaddle/Research/tree/master/
NLP/ACL2019-ARNOR} Besides, they also publish a test set for evaluating noise-filtering ability. Details of the adopted dataset are shown in Table \ref{tab:dataset}.


We construct the second part of experiments (Sec.\ref{noisy_tacred_expers}) to better understand SENT's behaviors. Since no labeled training data are available in the distant supervision setting, we construct a noisy dataset with 30\% noise from a labeled dataset, TACRED \cite{zhang-etal-2017-position} \footnote{https://github.com/yuhaozhang/tacred-relation}. We regard this constructed dataset as noisy-TACRED. The reason we choose this dataset is that 80\% instances in the training data are ``no\_relation". This ``NA" rate is similar to the NYT data which contains 70\% ``NA" relation type, thus analysis on this dataset is more credible. 

When constructing noisy-TACRED, the noisy instances are uniformly selected with 30\% noise ratio. Then, each noisy label is created by sampling a label from a complementary class with a weight of class frequency (in order to maintain the data distribution). Note that the original dataset consists of 80\% ``no\_relation" data, which means 80\% of the noisy instances are ``false-positive" instances, corresponding to the large amount of ``false-positive" noise in NYT-10. Details of the noisy-TACRED are also shown in Table \ref{tab:dataset}.




\setlength\extrarowheight{1.5pt}
\begin{table}[]
\begin{center}

\small
\begin{tabular}{|c|c|c|c|}
\cline{1-4}
\multicolumn{2}{|c|}{ Datasets } & {\bfseries NYT-10 } & {\bfseries noisy-TACRED } \\
\cline{1-4}
\multicolumn{2}{|c|}{\#Label num.}  & 24 &  41 \\
\cline{1-4}
\multirow{3}*{ Train}  & \#Instances & 371461 & 68124 \\
& \#Positive & 110518 & 26575  \\
& \#Noise & Unknown & 20586 \\

\cline{1-4}
\multirow{2}*{ Dev}  & \#Instances &2379 & 22631 \\
& \#Positive &337 & 5436\\
\cline{1-4}
\multirow{2}*{ Test}  & \#Instances & 2164 & 15509\\
& \#Positive & 323 &3325\\

\cline{1-4}

\end{tabular}
\end{center}
\caption{Statistics of datasets\footnotemark. ``Positive" means positive instances that are not labeled as ``NA". Note that the positive instances of noisy-TACRED include false-positive noise and the noise number in NYT-10 is unknown due to the inaccurate annotations. }
\label{tab:dataset}
\end{table}
\setlength\extrarowheight{0pt}


\subsection{Baselines}

We compare our SENT method with several strong baselines in distant RE. These compared methods can be categorized as: bag-level denoising methods, sentence-level denoising methods, sentence-level non-denoising methods.\footnotetext{Statistics of NYT-10 are quoted from \cite{jia-etal-2019-arnor}.}

\textbf{PCNN+SelATT} \cite{lin2016neural}: A \textit{bag-level} RE model which leverages an attention mechanism to reduce noise effect.

\textbf{PCNN+RA\_BAG\_ATT} \cite{ye2019distant} short for PCNN+ATT\_RA+BAG\_ATT, a \textit{bag-level} model containing both intra-bag and inter-bag attentions to alleviate noise.

\textbf{CNN+RL$_1$} \cite{qin2018robust}: A RL-based \textit{bag-level} method. Different from \textbf{CNN+RL$_2$}, they redistribute the filtered data into the negative examples.

\textbf{CNN+RL$_2$} \cite{feng2018reinforcement}: A \textit{sentence-level} RE model. It jointly train a instance selector and a CNN classifier using reinforcement learning (RL). 

\textbf{ARNOR} \cite{jia-etal-2019-arnor}: A \textit{sentence-level} RE model which selects confident instances based on the attention score on the selected patterns. It is the state-of-the-art method in sentence level.

\textbf{CNN} \cite{zeng-etal-2014-relation}, \textbf{PCNN} \cite{zeng2015distant} and \textbf{BiLSTM} \cite{zhang-etal-2015-bidirectional} are typical architectures used in RE.

\textbf{BiLSTM+ATT} \cite{zhang-etal-2017-position} leverages an attention mechanism based on BiLSTM to capture useful information. 

\textbf{BiLSTM+BERT} \cite{devlin-etal-2019-bert}: Based on BiLSTM, it utilizes the pre-trained BERT representations as word embedding. 

\begin{table*}[ht]\setlength{\tabcolsep}{2.6pt}
\centering
\myfontnew
\begin{tabular}{lccccccccc}
\toprule
\multirow{2}*{\bfseries Method} & \multicolumn{3}{c}{\bfseries Dev} & \multicolumn{3}{c}{\bfseries Test} \\
\cline{2-7}
& \textbf{Prec.}& \textbf{Rec.}& \textbf{F1}& \textbf{Prec.}& \textbf{Rec.}& \textbf{F1}\\

\midrule
CNN\cite{zeng-etal-2014-relation} & 38.32 & 65.22 & 48.28 & 35.75 & 64.54 & 46.01 \\
{PCNN}\cite{zeng2015distant} & 36.09 & 63.66 & 46.07 & 36.06 & 64.86 & 46.35 \\
{BiLSTM}\cite{zhang-etal-2015-bidirectional} & 36.71 & 66.46 & 47.29 & 35.52 & 67.41 & 46.53 \\
{BiLSTM+ATT}\cite{zhang-etal-2017-position} & 37.59 & 64.91 & 47.61 & 34.93 & 65.18 & 45.48 \\
{BERT}\cite{devlin-etal-2019-bert} & 34.78 & 65.17 & 45.35 & 36.19 & 70.44 & 47.81 \\
{BiLSTM+BERT}\cite{devlin-etal-2019-bert} & 36.09 & 73.17 & 48.34 & 33.23 & 72.70 & 45.61 \\
\midrule
{PCNN+SelATT}\cite{lin2016neural} & 46.01 & 30.43 & 36.64 & 45.41 & 30.03 & 36.15 \\
{PCNN+RA\_BAG\_ATT}\cite{ye2019distant} & 49.84 & 46.90 & 48.33 & 56.76 & 50.60 & 53.50 \\
{CNN+RL$_1$} \cite{qin2018robust} & 37.71 & 52.66 & 43.95 & 39.41 & 61.61 & 48.07 \\
{CNN+RL$_2$} \cite{feng2018reinforcement} & 40.00 & 59.17 & 47.73 & 40.23 & 63.78 & 49.34 \\
{ARNOR}\cite{jia-etal-2019-arnor} & 62.45 & 58.51 & 60.36 & 65.23 & 56.79 & 60.90 \\
\midrule
{\textbf{SENT (BiLSTM)}} & 66.71${_{\pm{0.30}}}$ & 57.27$_{\pm{0.30}}$ & 61.63$_{\pm{0.29}}$ & 71.22$_{\pm{0.58}}$ & 59.75$_{\pm{0.62}}$ & 64.99$_{\pm{0.34}}$ \\
{\textbf{SENT (BiLSTM+BERT)}} & \textbf{69.94$_{\pm{0.51}}$}& \textbf{63.11$_{\pm{0.61}}$} & \textbf{66.35$_{\pm{0.11}}$} & \textbf{76.34$_{\pm{0.56}}$} & \textbf{63.66$_{\pm{0.17}}$} & \textbf{69.42$_{\pm{0.13}}$} \\
\bottomrule

\end{tabular}
\caption{Main results on sentence-level evaluation. Compared baselines include normal RE model (the first part of the table) and models for distant RE (the second part of the table). We ran the model three times to get the average results.}
\label{tab:main}
\end{table*}

\subsection{Implementation Details}\label{implementation}
As SENT is a model-agnostic framework, we implement the classification model with two typical architectures: BiLSTM and BiLSTM+BERT. Since BiLSTM is also the base model of {ARNOR}, we can compare these two methods more fairly. During SENT training, we use the 50-dimension glove vectors as word embedding. While for PT after SENT, we randomly initialize the 50-dimension word embedding as the same in ARNOR. In both training phases, we use 50-dimension randomly-initialized position and entity type embedding. We train a single-layer BiLSTM with hidden size 256 using the adam optimizer at a learning rate of 5e-4. When implemented with BiLSTM+BERT, the setting is the same as those with BiLSTM except that we use a 768-dimension fixed BERT representation as word embedding (we use the ``bert-base-uncased" pre-trained model). We tune the hyperparameters on the development set via a grid search. Specifically, when training on the NYT dataset, we train the model for 10 epochs in each iteration, with the global data-filtering threshold $Th=0.25$, the re-labeling threshold $Th_{relabel}=0.7$ and negative samples number $K=10$. When training on the noisy-TACRED, we train for 50 epochs in each iteration, with $Th=0.15$, $Th_{relabel}=0.85$ and $K=50$. 

\textbf{To deal with the multi-label problem}, we utilize a simple method by randomly selecting one of the bag labels for each sentence. Such random selection turns the multi-label noise into the wrong-label noise, which is easier to handle. According to \citet{surdeanu-etal-2012-multi}, there are 31\% wrong-label noise and 7.5\% multi-label noise in NYT-10, and incorrect selection may result in 4\% extra wrong-label noise, which can be filtered out through NT identically with wrong-label instances.

\subsection{Sentence-Level Evaluation}
Table \ref{tab:main} shows the results of SENT and other baselines on sentence-level evaluation, where the results of SENT are obtained by PT after SENT. We can observe that: 1) Bag-level methods fail to perform well on sentence-level evaluation, indicating that it is difficult for these bag-level approaches to benefit downstream tasks with exact sentence labels. This result is consistent with the results in \citet{feng2018reinforcement}. 2) When performing sentence-level training on the noisy distant data, all baseline models show poor results, including the preeminent pre-trained language model BERT. These results indicate the negative impact of directly using bag-level labels for sentence-level training regardless of noise. 3) The proposed SENT method achieves a significant improvement over previous sentence-level de-noising methods. When implemented with BiLSTM, the model obtains a 4.09\% higher F1 score than ARNOR. Moreover, when implemented with BiLSTM+BERT, the F1 score is further improved by 8.52\%. 4) The SENT method achieves much higher precision than the previous de-noising methods when maintaining comparable or higher recall, indicating the effectiveness of the noise-filtering and re-labeling approaches.

\begin{table}[]
\centering
\myfont
\begin{tabular}{lccccc}
\toprule
\textbf{Noise Reduction} & \textbf{Prec.} & \textbf{Rec.} & \textbf{F1}  \\
\midrule
CNN+RL$_2$  & 40.58 & \textbf{96.31} & 57.10 \\
ARNOR  & 76.37 & 68.13 & 72.02\\
\textbf{SENT (BiLSTM)} & 80.00 & {88.46} & 84.02\\
\textbf{SENT (BiLSTM+BERT)} & \textbf{84.33} & {85.67} & \textbf{84.99}\\

\bottomrule

\end{tabular}
\caption{The noise-filtering effect evaluated on a noise-annotated test set of NYT-10.}
\label{tab:denoise}
\end{table}

\subsubsection{Noise-Filtering Effect on Distant Data}
In order to prove the effectiveness of SENT in de-noising distant data, we conduct a noise-filtering experiment following ARNOR. We use a test set published by ARNOR, which consists of 200 randomly selected sentences with an ``is\_noise" annotation. We perform a noise-filtering process as described in Sec.\ref{denoising}, and calculate the de-noise accuracy. As seen in Table \ref{tab:denoise}, the SENT method achieves remarkable improvement over ARNOR in F1 score by 12\%. While improving in precision, SENT achieves 20\% gain over ARNOR in recall. As ARNOR initializes the training data with a small part of frequent patterns, these patterns might limit the model from generalizing to various correct data. Different from ARNOR, SENT leverages negative training to automatically learn the correct patterns, showing better ability in diversity and generalization.

\setlength\extrarowheight{2pt}

\begin{table}[]\setlength{\tabcolsep}{2pt}
\centering
\myfont
\begin{tabular}{c|lcccc}
\toprule
& \textbf{Method} & \textbf{Prec.} & \textbf{Rec.} & \textbf{F1}  \\
\cline{1-5}

{Clean} & BiLSTM+ATT  & 67.7 & 63.2 & 65.4 \\
{Data} & BiLSTM  & 61.4 & 61.7 & 61.5\\
\cline{1-5}

\multirow{3}*{\makecell[l]{Noisy \\ Data}} & BiLSTM+ATT  & 32.8 & 43.8 & 37.5 \\
 & BiLSTM  & 37.8 & 45.5 & 41.3\\
& \textbf{SENT (BiLSTM)} & \textbf{66.0} & \textbf{52.9} & \textbf{58.7}\\

\bottomrule

\end{tabular}
\caption{Model performance on clean and noisy-TACRED. When trained on noisy data, the performance of base models degrades dramatically while SENT achieves comparable results with the models trained on clean data.}
\label{tab:tacred_denoise}
\end{table}

\setlength\extrarowheight{0pt}

\begin{figure}
    \centering
    \includegraphics[width=1.0\linewidth]{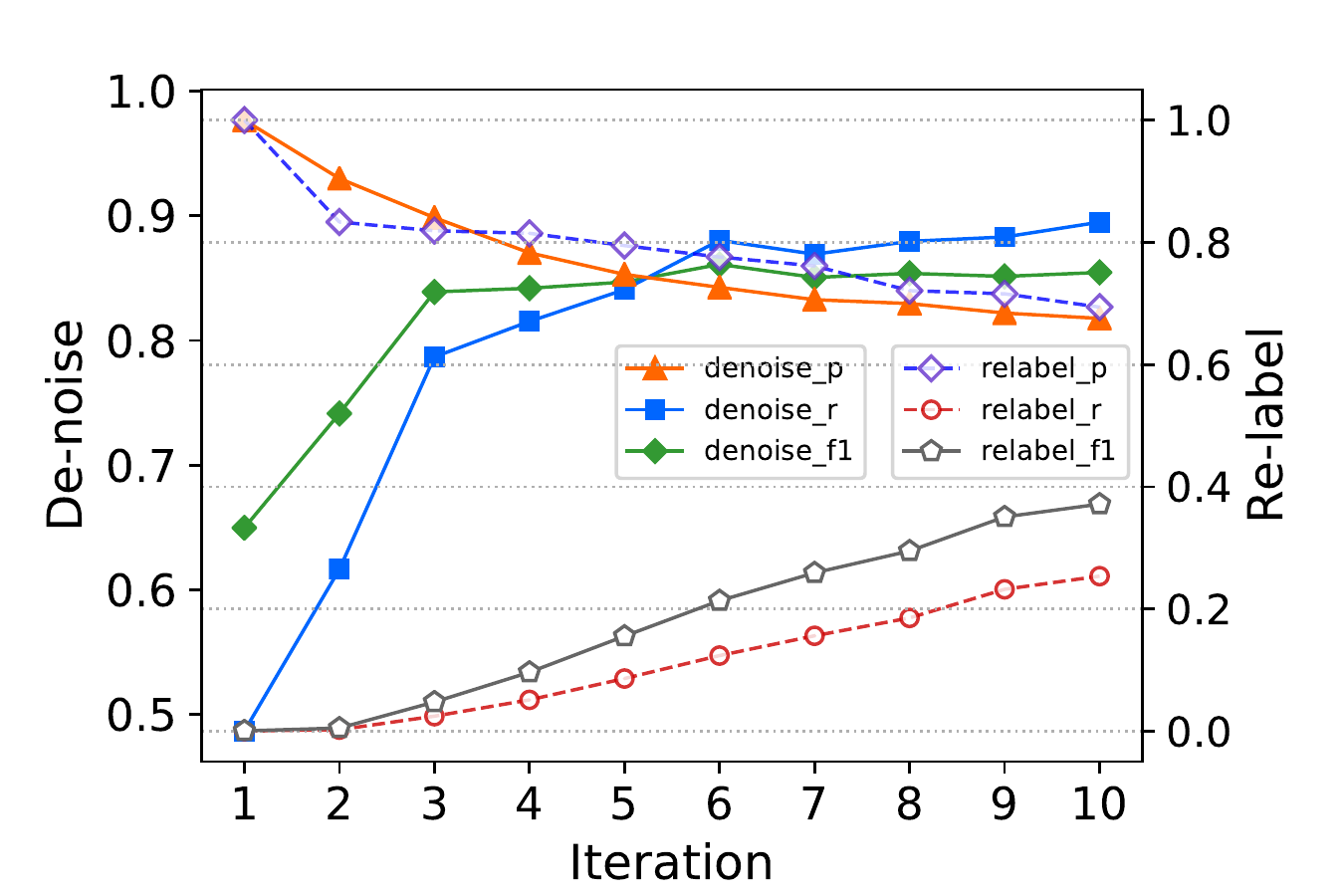}
    \caption{Data-refining details on noisy-TACRED.}
    \label{fig:data_refine}
\end{figure}

\subsection{Analysing SENT on ``Labeled Noise"} \label{noisy_tacred_expers}
In this section, we analyze the effectiveness of the data-refining process with a self-constructed noisy data set: noisy-TACRED (details in Table \ref{tab:dataset}).


\subsubsection{Performance on Noisy-TACRED}
Table \ref{tab:tacred_denoise} shows the results of training on TACRED and noisy-TACRED. As seen, the baseline model degrades dramatically on the noisy data, with the LSTM dropping by 20.2\%. However, after training with SENT, the BiLSTM model can achieve comparable results with the model that trained on the clean data. Note that the de-noising method is quite helpful in promoting the precision score, yet the recall is still lower than that on clean data.

\subsubsection{Effects of Data-Refining}
We also evaluate the noise-filtering and label-recovering ability on the noisy-TACRED training set, as shown in Fig.\ref{fig:data_refine}. We can observe that: 1) SENT achieves about 85\% F1 score on the noisy-TACRED data. This result is consistent with the noise-filtering results obtained on the NYT dataset (with 200 sampled instances), validating the de-noising ability of SENT on different datasets. 
2) As the training iteration progressed, the precision of noise-filtering decreases with the recall promoting.
More noise-filtering contributes to a cleaner dataset, while it might bring more false-noise mistakes. Therefore, we stop the iteration when the model reaches the best score on the development set.
3) As for label-recovering, SENT can achieve about 70\% precision with about 25\% recall. Here, the threshold setting is also a trade-off that we prefer to adopt a modest value for more accurate re-labeling.

\setlength\arrayrulewidth{0.2pt}
\setlength\extrarowheight{2pt}

\begin{table*}[]\setlength{\tabcolsep}{7pt}\scriptsize
    \centering
    \begin{tabular}{|c|c|c|c|lll}
        \cline{1-4}
        \textbf{Sentence}&\textbf{Bag label}&\textbf{Sentence label}&\textbf{Refined label} \\
        \cline{1-4}

        \makecell[l]{The plan filed on behalf of the state 's Democratic Congressional delegation , for instance
        , would \\make the 25th district , which zigzags 300 miles from southern \textcolor{blue}{Austin} to Mexico , 
        much shorter\\ and Austin-based , which would help the incumbent Democrat , \textcolor{red}{Lloyd Doggett} . } & \multirow{4}*{\makecell[l]{place\_lived \\ place\_of\_birth}} & place\_lived & NA 
        \\ 
        \cline{1-1} \cline{3-4}
        \makecell[l]{It would draw the lines in a way that imperils an incumbent Democrat , Representative  \textcolor{red}{Lloyd} \\ \textcolor{red}{Doggett} of \textcolor{blue}{Austin} , and divides that most liberal of Texas cities and surrounding Travis County\\ among three districts , all solidly Republican . ''  } & & place\_of\_birth & place\_lived 
        \\ 
 
        \cline{1-4}

        \makecell[l]{A leather-and-metal chair bore a shameless resemblance to a Barcelona Chair by \textcolor{red}{Ludwig Mies } \\ \textcolor{red}{van der Rohe} -LRB- who lived in \textcolor{blue}{Chicago} -RRB- . } & \multirow{5}*{place\_of\_death} & place\_of\_death & NA
        \\ 
        \cline{1-1} \cline{3-4}
        \makecell[l]{ The works of architects like Frank Lloyd Wright , Louis H. Sullivan , \textcolor{red}{Ludwig Mies van der Rohe}\\ and Helmut Jahn define \textcolor{blue}{Chicago} in many ways . } & & place\_of\_death & NA 
        \\ 
        \cline{1-1} \cline{3-4}
        \makecell[l]{Mr. Freed received a bachelor 's degree in architecture in 1953 from the Illinois Institute of Tech- \\nology in \textcolor{blue}{Chicago} , which was then under the direction of Ludwig \textcolor{red}{Ludwig Mies van der Rohe} . } & & place\_of\_death & NA
        \\ 
 
        \cline{1-4}

        \makecell[l]{It 's really tough right now , '' said \textcolor{red}{Norman J. Ornstein} , a resident scholar at the conservative\\ \textcolor{blue}{American Enterprise Institute} and a member of the PBS board . '' } & {NA} & NA & company
                \\ 
        \cline{1-1} \cline{2-4}
        \makecell[l]{Three of the sailors were assigned to SEAL Delivery Team 1 , \textcolor{blue}{Pearl Harbor} , \textcolor{red}{Hawaii} . } & NA & NA & contains
        \\ 
        \cline{1-1} \cline{2-4}
        \makecell[l]{ An obituary on Wednesday about \textcolor{red}{Philip Merrill} , a \textcolor{blue}{Maryland} publisher , misstated a journalism \\post he held as an undergraduate .  } & NA & NA & place\_lived

        \\ 
 
        \cline{1-4}

    \end{tabular}
    \caption{Examples showing the ability of SENT to refine the bag-level noisy data into correct data. Texts in red and blue denote the head and tail entity, respectively.
    }

    \label{tab:samples}
\end{table*}





\begin{figure}[htbp]
\centering
\subfigure[Head relation.]{
\begin{minipage}[t]{0.5\linewidth}
\includegraphics[width=1.0\linewidth]{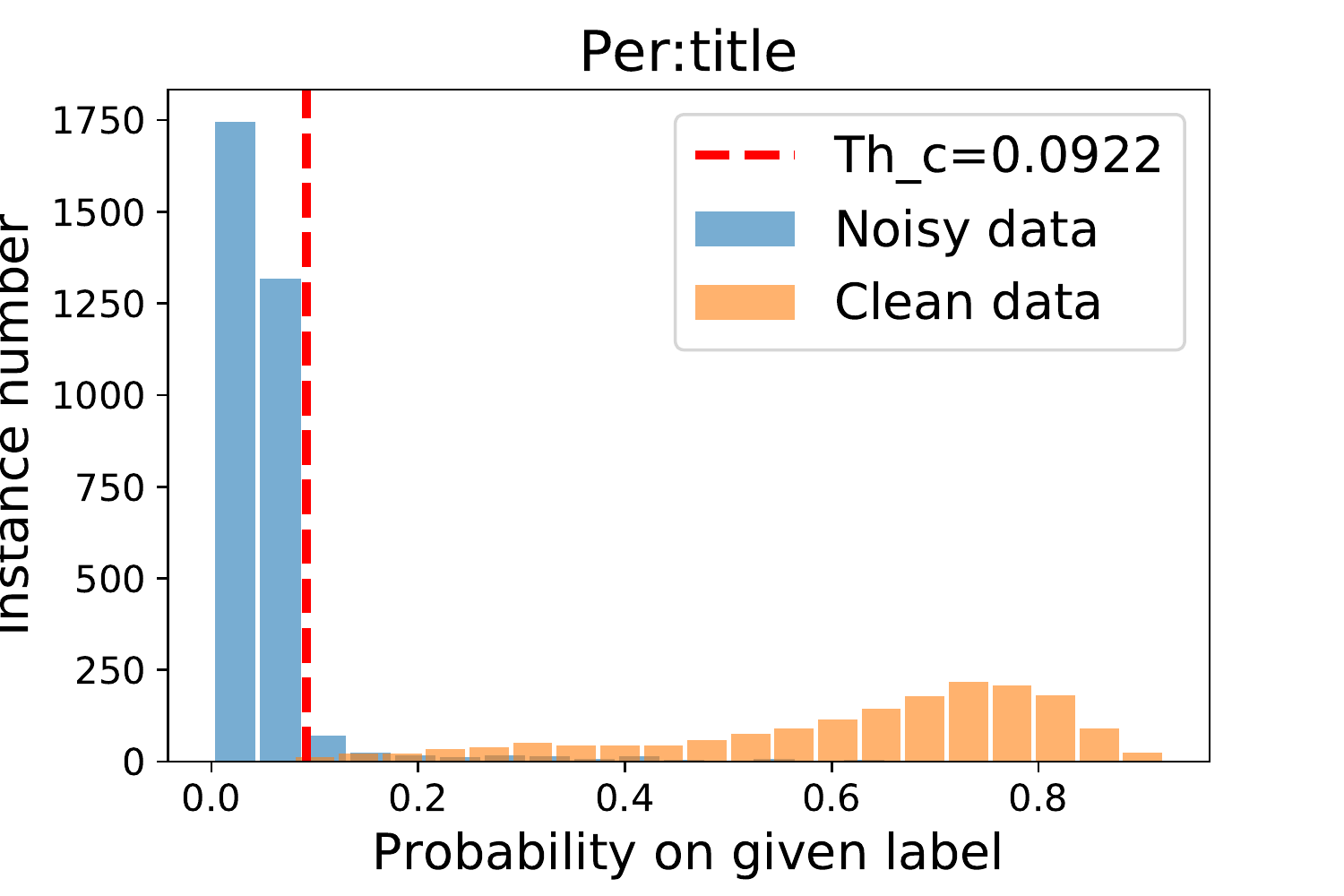}
\end{minipage}%
}%
\subfigure[Long-tail relation.]{
\begin{minipage}[t]{0.5\linewidth}
\includegraphics[width=1.0\linewidth]{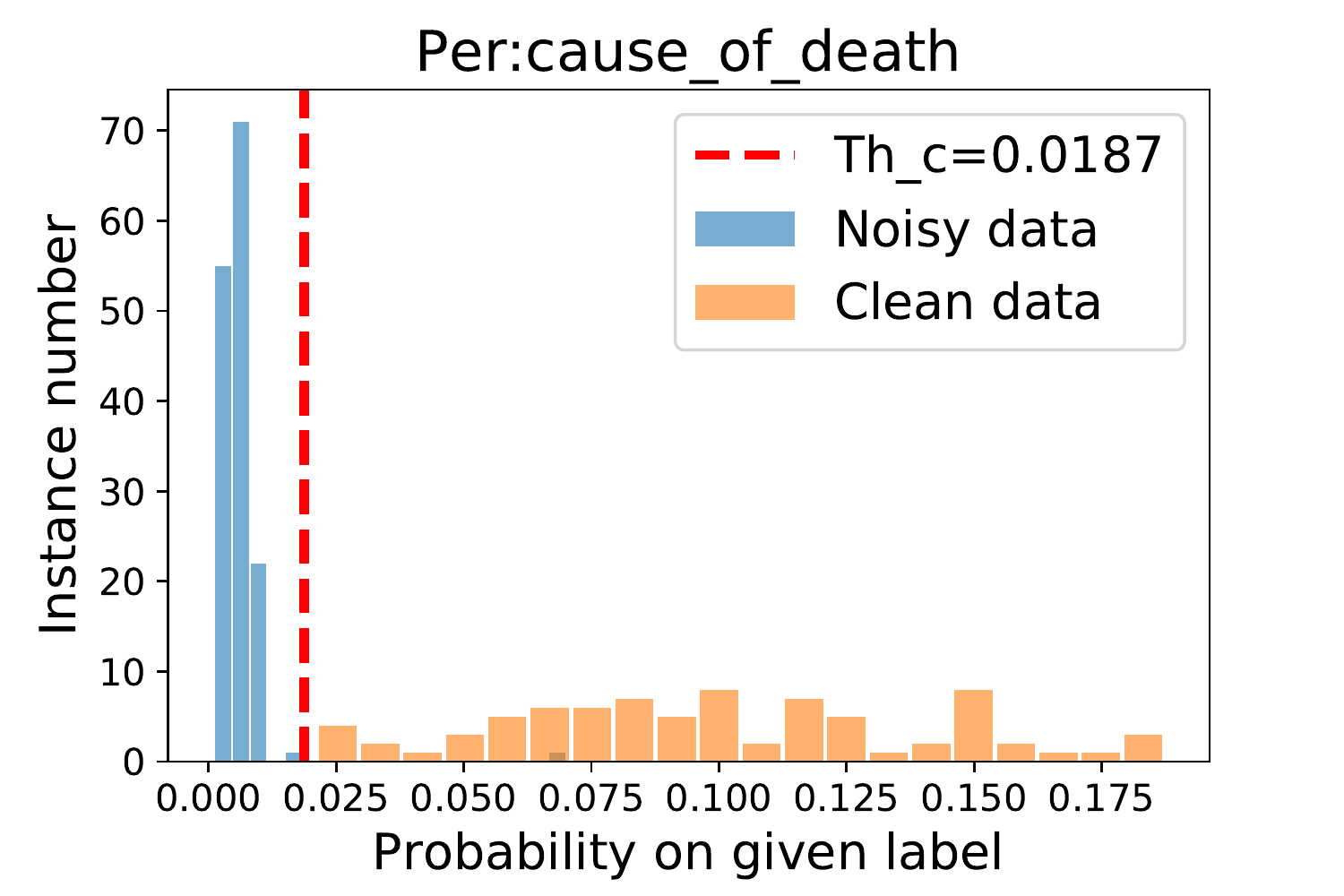}
\end{minipage}%
}%

\centering
\caption{Data distribution of a head relation (per:title) and a long-tail relation (per:cause\_of\_death) during NT. The dynamically designed thresholds benefits filtering.}
\label{fig:head_tail}
\end{figure}

\subsubsection{Effects of Dynamically Filtering}
As described in Sec.\ref{filter_relabel}, we design a dynamic filtering threshold for long-tail data. The effect of this strategy is shown in Fig.\ref{fig:head_tail}. As seen, the degree of convergence of the long-tail relation ``per:cause\_of\_death" is much lower than that of the head relation. Simply setting a uniform threshold would harm the data distribution with instances of ``per:cause\_of\_death" largely filtered. While with a dynamically determined threshold, both data from the head and the long-tail relations are appropriately filtered.

\begin{table}[]
\centering
\myfont
\begin{tabular}{lccccc}
\toprule
\textbf{Components} & \textbf{Prec.} & \textbf{Rec.} & \textbf{F1}  \\
\midrule
SENT (BiLSTM)  & 71.22 & \textbf{59.75} & \textbf{64.99} \\
$-$ Final PT  & \textbf{72.48} & 57.89 & 64.37\\
{$-$ Re-labeling} & 66.67 & {55.11} & 60.34\\
{$-$ Dynamic threshold} & {58.46} & {49.23} & {53.45}\\
{$-$ Re-initialization} & {48.61} & {65.02} & {55.63}\\
{$-$ NT} & {41.58} & {70.28} & {52.24}\\

\bottomrule

\end{tabular}
\caption{An ablation study on NYT-10. }
\label{tab:ablation}
\end{table}

\subsection{Ablation Study}
To better illustrate the contribution of each component in SENT, we conduct an ablation study by removing the following components: final PT, re-labeling, dynamic threshold, re-initialization, NT. The test results are shown in Table \ref{tab:ablation}. We can observe that: 1) Removing the final positive training affects little to the performance. This is because the model trained with NT already reaches high accuracy and the purpose of final PT is only to achieve more confidential predictions. 2) Removing the re-labeling process harms the performance, as the filtered instances are simply discarded regardless of the useful information for training. 3) Without dynamic threshold, clean instances from the tail classes are incorrectly filtered out, which severely degrades the performance. 4) Re-initialization also contributes a lot to the performance. The model trained on the original noisy data inevitably fits to the noisy distribution, while re-initialization helps wash out the overfitted parameters and eliminate the noise effects, thus contributing to better training and noise-filtering. 5) Training with PT instead of NT causes a dramatic decline in performance, especially on the precision, which verifies the effectiveness of NT to prevent the model from overfitting noisy data.

\subsection{Case Study}
As discussed, SENT is able to refine the distant RE dataset. In fact, there exists much noise in the NYT data that is difficult to tackle with bag-level methods. In Table \ref{tab:samples}, we show some examples. (1) The first two rows are the sentences in a multi-label bag. We randomly choose one of the bag labels for each sentence, and the model is able to correct the bad choice (by correcting the second sentence with ``place\_lived" and the first sentence with ``NA"). (2) The following three rows show a bag with label ``place\_of\_death", while this whole bag is actually a ``NA" bag incorrectly labeled positive. (3) SENT can also recognize the positive samples in ``NA". As shown in the last three rows, each sentence labeled as ``NA" is actually expressing a positive label. In fact, such false-negative problem is frequently seen in the NYT data, which contains 70\% negative instances that were labeled ``NA" only because the entity pairs do not participate in a relation in the database. We believe the capacity to recognize these false-negative samples can significantly boost the performance.

\section{Conclusion}

In this paper, we present SENT, a novel sentence-level framework based on Negative Training (NT) for sentence-level training on distant RE data. 
NT not only prevent the model from overfitting noisy data, but also separate the noisy data from the training data. By iteratively performing noise-filtering and re-labeling based on NT, SENT helps re-fine the noisy distant data and achieves remarkable performance. Experimental results verify the improvement of SENT over previous methods on sentence-level relation extraction and noise-filtering effect.

\section*{Acknowledgements}
The authors wish to thank the anonymous reviewers for their helpful comments. This work was partially funded by China National Key R\&D Program (No. 2018YFB1005100), National Natural Science Foundation of China (No. 61976056, 62076069), Shanghai Municipal Science and Technology Major Project (No.2021SHZDZX0103).
\bibliographystyle{acl_natbib}
\bibliography{anthology,acl2021}


\end{document}